\documentclass[10pt,twocolumn,letterpaper]{article}

\usepackage{iccv}              
\usepackage{flushend}

\usepackage{adjustbox}
\usepackage{array}
\usepackage{float}
\usepackage{makecell}
\usepackage{xr-hyper} 
\usepackage{multirow}
\usepackage[accsupp]{axessibility}  

\definecolor{iccvblue}{rgb}{0.21,0.49,0.74}
\usepackage[pagebackref,breaklinks,colorlinks,allcolors=iccvblue]{hyperref}

\def\paperID{*****} 
\def\confName{ICCV}
\def\confYear{2025}

\title{CytoDiff: AI-Driven Cytomorphology Image Synthesis for Medical Diagnostics}

\author{
Jan Carreras Boada$^{1,4,\ast}$, Rao Muhammad Umer$^{1,\ast}$, Carsten Marr$^{1,2,3}$\\
$^{1}$ Institute of AI for Health, Helmholtz Zentrum München - German Research Center \\ 
for Environmental Health, Neuherberg 85764, Germany\\
$^{2}$ Department of Medicine III, Ludwig-Maximilian-University Hospital, Munich, Germany\\
$^{3}$ DKTK, German Cancer Consortium, Germany\\
$^{4}$ Escola Superior de Comerç Internacional, Universitat Pompeu Fabra (ESCI-UPF),\\
Barcelona 08003, Spain \\
\tt\small{carsten.marr@helmholtz-munich.de}
}

\begin{document}
\maketitle

\renewcommand{\thefootnote}{\fnsymbol{footnote}}
\footnotetext[1]{Contributed equally}
\renewcommand{\thefootnote}{\arabic{footnote}}

\begin{abstract}
Biomedical datasets are often constrained by stringent privacy requirements and frequently suffer from severe class imbalance. These two aspects hinder the development of accurate machine learning models. While generative AI offers a promising solution, producing synthetic images of sufficient quality for training robust classifiers remains challenging.
This work addresses the classification of individual white blood cells, a critical task in diagnosing hematological malignancies such as acute myeloid leukemia (AML). We introduce CytoDiff, a stable diffusion model fine-tuned with LoRA weights and guided by few-shot samples that generates high-fidelity synthetic white blood cell images. Our approach demonstrates substantial improvements in classifier performance when training data is limited. Using a small, highly imbalanced real dataset, the addition of 5,000 synthetic images per class improved ResNet classifier accuracy from 27\% to 78\% (+51\%). Similarly, CLIP-based classification accuracy increased from 62\% to 77\% (+15\%). These results establish synthetic image generation as a valuable tool for biomedical machine learning, enhancing data coverage and facilitating secure data sharing while preserving patient privacy. Paper code is publicly available at \url{https://github.com/JanCarreras24/CytoDiff}.
\end{abstract}    
\section{Introduction}
\label{sec:intro}
Accurate classification of white blood cells is essential to support clinical decisions and detect hematological disorders \cite{walter2022artificial}. One current challenge in the classification of cytomorphological images is the lack of sufficient real data to effectively train models for rare classes \cite{umer2023imbalanced}. It is widely acknowledged that the efficacy of a classifier is contingent on the quality and quantity of the training dataset \cite{banerjee2023machine}. In many cases, however obtaining an optimal dataset is not feasible, particularly when dealing with underrepresented conditions or infrequent cell types, where gathering sufficient data is expensive and time-consuming \cite{ye2023synaug}.

Furthermore, managing patient data often requires strict privacy and authorization considerations,  which can ultimately hinder access and collaboration across institutions \cite{rieke2020future, sun2025data}. An alternative is offered by synthetic data, which enables data availability and facilitates collaborative research without compromising patient confidentiality \cite{pezoulas2024synthetic}. Thus, high-quality synthetic images could serve as a reliable complement to real-world medical data \cite{coyner2022synthetic}. 

\begin{figure}[t!]
    \centering
    \begin{adjustbox}{width=0.48\textwidth}
    \setlength{\tabcolsep}{4pt}
    \renewcommand{\arraystretch}{1.2}

    \newcommand{\bigcell}[1]{%
      \begin{minipage}[c]{2.5cm} 
      \centering
      \fontsize{15}{18}\selectfont
      #1
      \end{minipage}%
    }

    \begin{tabular}{cccccc}
        \raisebox{0.3cm}{\rotatebox{90}{\bigcell{Real}}} &
        \includegraphics[width=0.18\textwidth]{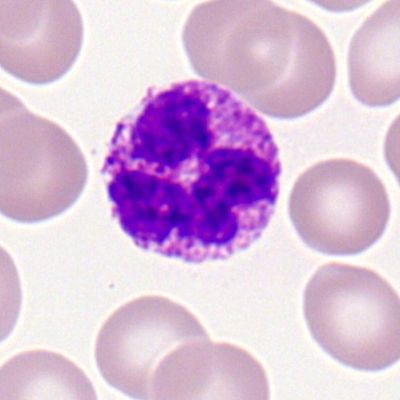} &
        \includegraphics[width=0.18\textwidth]{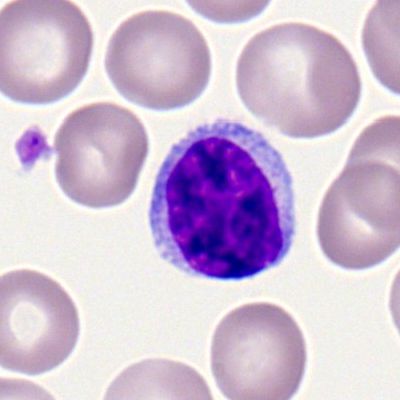} &
        \includegraphics[width=0.18\textwidth]{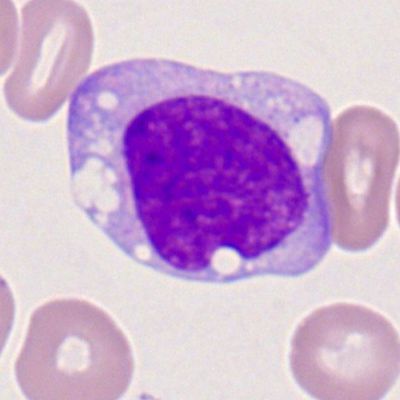} &
        \includegraphics[width=0.18\textwidth]{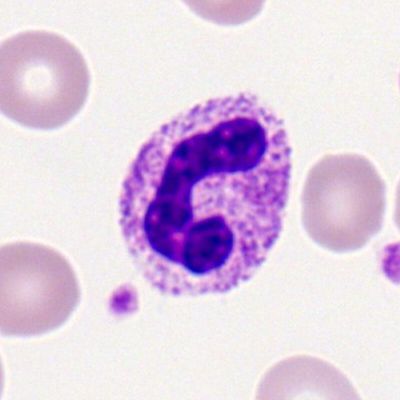} &
        \includegraphics[width=0.18\textwidth]{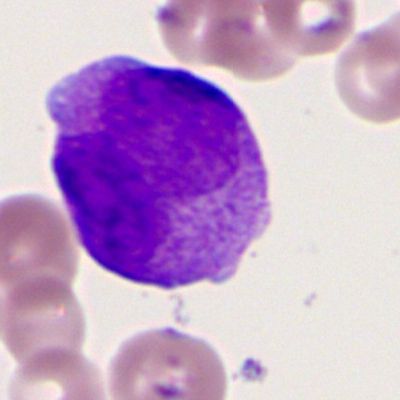} \\
        
        \raisebox{0.4cm}{\rotatebox{90}{\bigcell{Synthetic}}} &
        \includegraphics[width=0.18\textwidth]{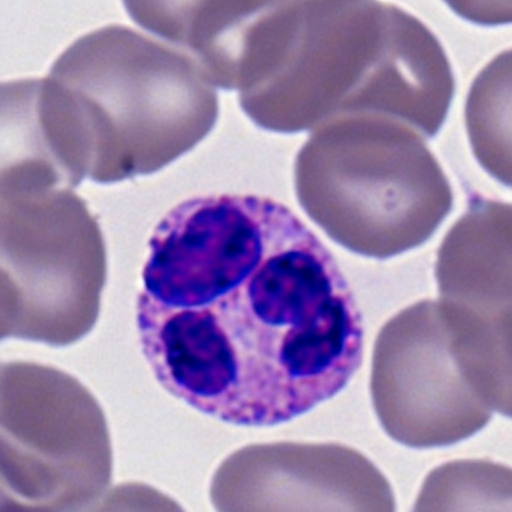} &
        \includegraphics[width=0.18\textwidth]{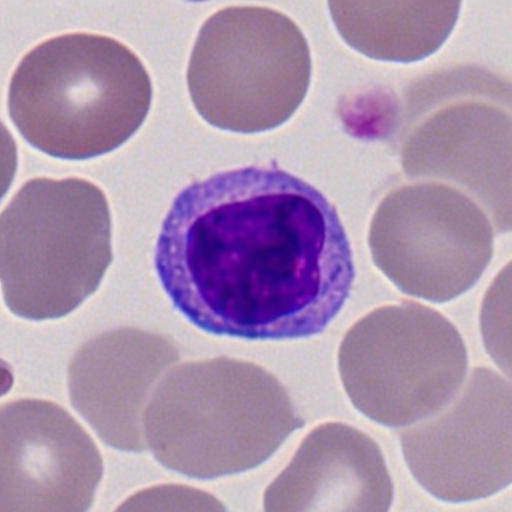} &
        \includegraphics[width=0.18\textwidth]{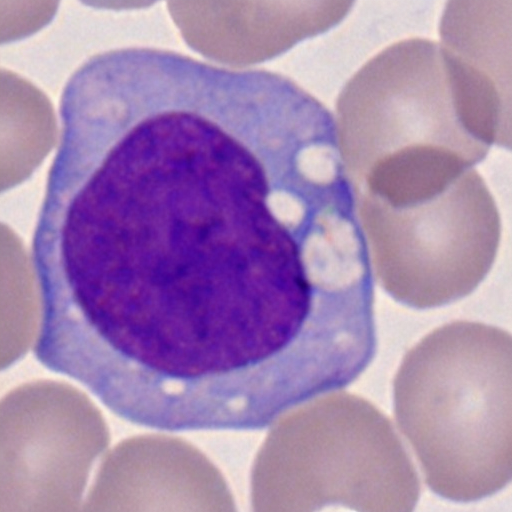} &
        \includegraphics[width=0.18\textwidth]{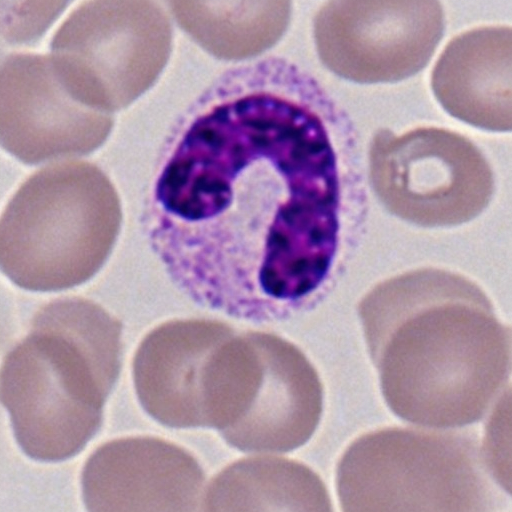} &
        \includegraphics[width=0.18\textwidth]{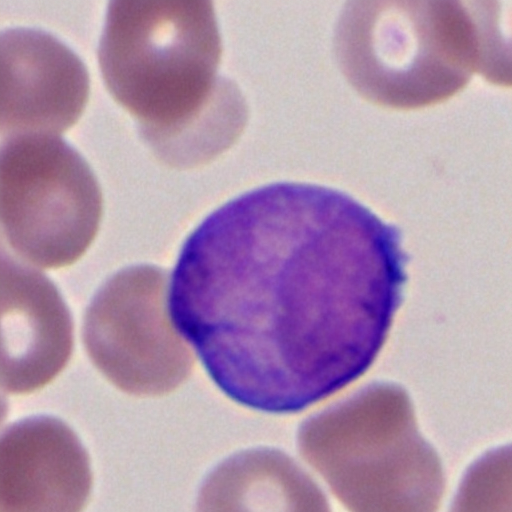} \\
        
        & \hspace{0.8mm}\bigcell{Basophil} & 
          \hspace{-1.3mm}\bigcell{Typical \\ lymphocyte} & 
          \hspace{-1mm}\bigcell{Monoblast} & 
          \hspace{0.2mm}\bigcell{Band \\ neutrophil} & 
          \hspace{-4.3mm}\bigcell{Promyelocyte \\ \makebox[0pt][l]{\hspace{-5.4mm}bilobed}}

    \end{tabular}

    \end{adjustbox}
    \caption{Real images (top) and synthetic images (bottom) generated by CytoDiff for five white blood cell classes.}
    \label{fig:comparison_real_synthetic}
    \vspace{-0.3cm}
\end{figure}

In recent advancements of text-to-image generative models~\cite{rombach2022high, podell2023sdxl, betker2023improving, saharia2022photorealistic, ruiz2023dreambooth}, diffusion models have emerged as a superior alternative to Generative Adversarial Networks (GANs)~\cite{goodfellow2014generative} in the domain of synthetic image generation. The capacity to produce high-quality images that accurately reflect the distinctive morphological characteristics of cellular structures renders them well-suited for medical applications \cite{bhosale2025pathdiff,susladkar2025victr}. Furthermore, by effectively using diffusion models, it is possible to generate synthetic data that preserve the characteristic visual appearance and unique biological features of each cell type \cite{bhosale2025pathdiff}. This opens the door to the possibility of obtaining unlimited data that reflect the morphological aspects of cells without privacy restrictions, potentially representing a major breakthrough in the biomedical field \cite{sheng2025synthetic}.

Despite recent advances, limitations still remain in cytomorphology, particularly regarding the generation of fine details for rare white blood cell classes with only few real samples. We here introduce a Cytomorphology Diffusion (CytoDiff) model for generating single white blood cells. It uses a few real samples allow to train a model capable of generating a large amount of synthetic data. Instead of applying traditional image augmentations, i.e., rotations, brightness, or color-jittering, CytoDiff is capable of generating new data that is not just a tuned copy of the originals, but novel instances that preserve biological plausibility, as illustrated in the comparison between real and synthetic single white blood cell images in Figure~\ref{fig:comparison_real_synthetic}.

After synthetic image generation of single white blood cells, we train the classifier with high imbalanced real data, balanced synthetic data, and combination of real and synthetic data to show the effectiveness of CytoDiff model. To summarize, the contributions of our work are as follows:

\begin{itemize}[label=\textbullet]
    \item We introduce CytoDiff, a stable diffusion model for generating highly accurate single white blood cell images.
    \item Under real few-shot guidance and customized text prompts with LoRA weights fine-tuning, the generated synthetic data align the best with the real images of rare classes by CytoDiff model (refer to section \ref{sec:few-shot ablation study}).
    \item We extensively evaluate the image generation and classification pipeline to validate the effectiveness of our CytoDiff model (refer to sections \ref{sec:Quantitative results comparison} and \ref{sec:Ablation study}).
\end{itemize}

\section{Methodology}


\subsection{Single white blood cells image generation}
Our goal is to improve classifier performance when trained on both real and synthetic data. In this work, the generative model stable diffusion 2.1~\cite{rombach2022high} is used, a diffusion model trained on millions of natural images, capable of generating high-quality images from random noise. Once trained, it can receive a text description (prompt) and generate an image related to the prompt. 

The choice of stable diffusion 2.1 as the foundation for synthetic image generation is motivated by two factors: (i) the denoising-based diffusion architecture ensures high-resolution and stable image synthesis, which is crucial in the biomedical domain. (ii) Stable diffusion is open-source and allows for extensive customization, including the integration of Low-Rank Adaptation (LoRA) weights, which enable efficient fine-tuning for domain-specific tasks. This flexibility proved particularly beneficial as it allowed us to specify detailed morphological characteristics directly in the text prompts. By guiding the generation process through carefully crafted descriptions, we are able to synthesize single white blood cell images that closely resemble their real counterparts in structure and appearance, which is essential for their effective use in downstream classification tasks.

While stable diffusion 2.1 is capable of generating high-quality images, it frequently exhibits a lack of precision when attempting to produce a specific image, such as a white blood cell out of the box. The result is devoid of realism and the morphological characteristics necessary for use in diagnostic classification, as can be observed in Supplementary Figure~S1.

\begin{figure}[t!]
    \centering
    \includegraphics[width=\linewidth]{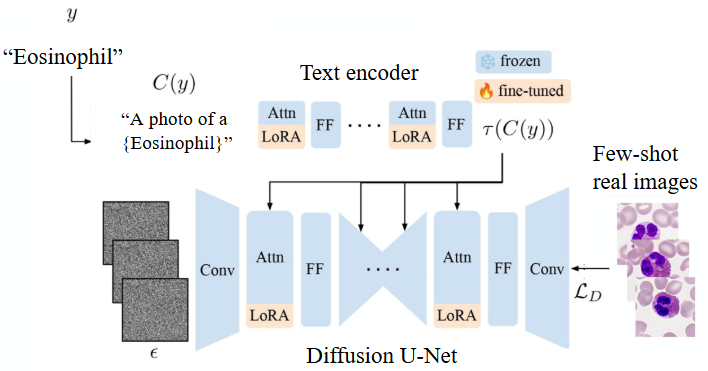}
    \caption{Overview of CytoDiff training. We fine-tune LoRA weights with modified stable diffusion 2.1 model for the linear weights of the attention layers in both the text-encoder and the diffusion U-Net encoder. We provide few-shot real images and prompts to better learn fine-grained details of each classes to generate images closer to the target class.
    Abbreviations: Attn = Attention layer, FF = Feed-forward layer, Conv = Convolutional layer.}
    \label{fig:workflow_col_fix}
    \vspace{-0.3cm}

\end{figure}

Inspired by the DataDream method \cite{kim2024datadream}, the utilization of LoRA weights enables the implementation of minor yet effective alterations to stable diffusion 2.1 \cite{choi2024simple}. The efficacy of this approach is predicated on the incorporation of two low-rank matrices within the attention layers of the text encoder and the Diffusion U-Net image encoder (Figure~\ref{fig:workflow_col_fix}). The modification facilitates the model to prioritize the most pertinent components of the prompt, thereby obviating the necessity for retraining of all layers. Our CytoDiff model integrates the text-to-image generation facilitated by stable diffusion 2.1, and image-to-image generation enabled by the trained LoRA weights, as illustrated in Figure~\ref{fig:workflow_col_fix}.


The initial step in the image generation process is to train the LoRA weights using real samples. Four different experiments are conducted per class, each using a different number of images: 1, 4, 8, and 16. It is very important that these images are as sharp as possible. To ensure this, a manual selection process is carried out to choose 16 images per class that clearly display the biological characteristics, since the used dataset contains a wide range of image quality (see Figure S2 of the supplementary material).

Subsequently, specific prompts are written for each class. These prompts not only specify the type of blood cells but also include essential details that each cell type must exhibit. For each class, a trial-and-error study is conducted to select the best prompts. An example of a final prompt is the following:

\begin{quote}
``Photorealistic basophil under microscope, peripheral blood smear, large bilobed nucleus, dark blue-purple granules densely packed in cytoplasm, surrounded by red blood cells, medical cytology, high detail, clinical pathology, Wright-Giemsa stain, white background, soft lighting, 40x magnification, ultra-detailed, sharp focus, macro lens.''
\end{quote}

\subsection{Single white blood cells image classification}

Subsequent to the generation of synthetic images, various experiments are conducted by using two different classifiers: ResNet-50~\cite{he2016deep} and CLIP~\cite{radford2021learning}. Each classifier has been trained in different experiments: including the use of solely real images, solely synthetic images, and a combination of both real and synthetic images. This setup enables the evaluation of several aspects. By training only with real data, we can assess whether classifiers are capable of achieving good performance despite strong class imbalance, and establish a baseline for addressing the research question: demonstrate whether synthetic images, generated using few real samples, can improve the performance of classification models and support medical diagnosis. Training solely on synthetic images allows us to evaluate whether classifiers can recognize the morphological features present in synthetic samples. Finally, the mixed dataset scenario makes it possible to analyze whether the inclusion of synthetic data is beneficial in biomedical applications.

ResNet-50~\cite{he2016deep} is a convolutional neural network (CNN) with 50 layers and residual connections. It is widely used in computer vision tasks due to its ability to learn deep hierarchical features while avoiding vanishing gradients. In this work, ResNet-50 is adapted for multi-class classification by replacing its final fully connected layer with a layer matching the number of cell types. The architecture is used to learn morphological features from real and synthetic cell images in a purely supervised setting.

CLIP (Contrastive Language–Image Pretraining)~\cite{radford2021learning}, on the other hand, is a dual-encoder model that jointly learns representations of images and text. In this work, the image encoder is ViT-B/32 variant, a Vision Transformer model, while the text encoder is based on a standard transformer architecture. Unlike traditional classifiers, CLIP aligns images and textual prompts in a shared embedding space. During the experiments, each test image is compared to the embeddings of the class-specific prompts, and classification is performed by selecting the class with the highest similarity. This architecture is particularly relevant for evaluating the consistency between the visual content of the image and the semantic information provided by the prompts.

By comparing both architectures under different data regimes; real only, synthetic only, and mixed, the study aims to assess the generalizability of synthetic data and its integration into traditional and prompt-based classifiers.

The primary loss function used in all experiments is cross-entropy loss, a standard choice for multi-class classification task. It quantifies the difference between the predicted class probabilities and the true labels, encouraging the model to assign high probability to the correct class.

When using only real or only synthetic images, the cross-entropy loss is computed over the entire training dataset in the standard way. In both cases, cross-entropy is used to optimize the models’ predictions, and AdamW was employed as the optimizer to adjust the weights through backpropagation. In mixed experiments, where both real and synthetic images are used, a combined loss strategy is implemented to control the contribution of each image type. A study is conducted to evaluate the best way to define the loss function, considering the different nature of the images. One of the strategies evaluated is outlined below:
\[
\text{Loss}_{\text{total}} = \lambda_1 \cdot \left( \frac{n_{\text{real}}}{n} \cdot \text{Loss}_{\text{real}} \right) + (1 - \lambda_1) \cdot \left( \frac{n_{\text{synth}}}{n} \cdot \text{Loss}_{\text{synth}} \right)
\]

where $n = n_\text{real} + n_\text{synth}$  and $\lambda_1 \in [0,1]$. This formulation allows adjusting the relative importance attributed to each type of image. In the final strategy employed, the loss function is computed in a manner that treated both image types equally.

\section{Experimental Results}

\subsection{Dataset details}
\begin{figure}[t!]
    \centering
    \includegraphics[width=\linewidth]{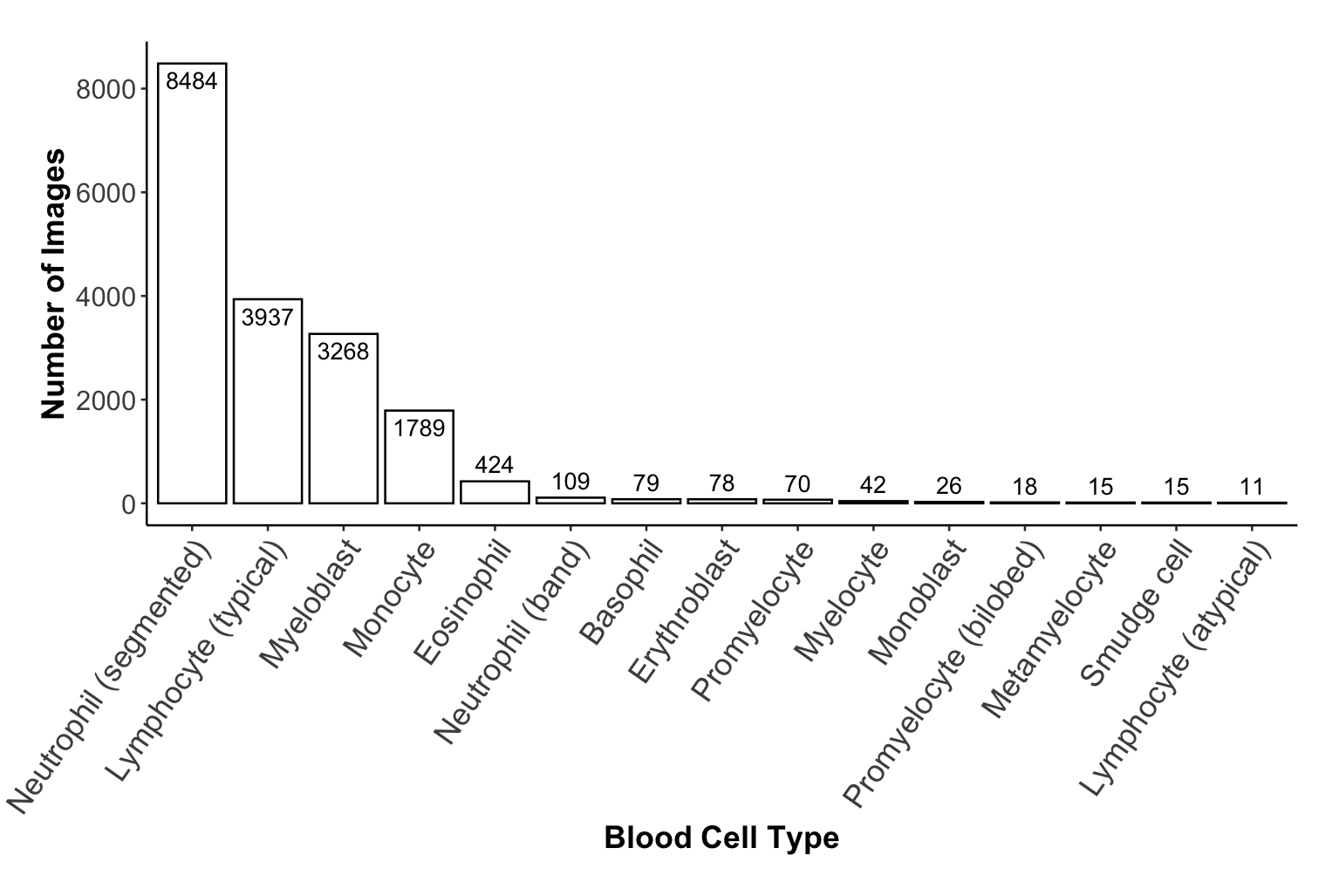}
    \caption{Class distribution of samples in the Munich AML Morphology Dataset.}
    \label{fig:dataset_col}
\end{figure}

The Munich AML Morphology Dataset \cite{matek2019dataset} includes 18,365 expert-labeled single-cell images.
The images are from peripheral blood smears of 100 patients without hematological malignancy and 100 patients diagnosed with acute myeloid leukemia. The patients were treated at Munich University Hospital between 2014 and 2017. The images were captured using the M8 digital microscope/scanner at 100x magnification with oil immersion, ensuring excellent resolution and distinct cell morphology. Pathological and non-pathological leukocytes were classified into a standard morphological scheme derived from clinical practice by trained experts. The dataset contains 15 single white blood cell classes with highly imbalanced sample counts, ranging from thousands of images in some classes to fewer than twenty in others (Figure~\ref{fig:dataset_col}). This imbalance provides an ideal scenario to evaluate whether our approach improves classifier performance, especially for minority classes.

All experiments are subjected to an essential step in the data, applying a 5-fold cross-validation to ensure consistency and robustness in the results. For each of the folds, the data has been split as follows; 60\% for the training set, 20\% for the validation after each epoch, and 20\% for the test set, to test the model with unseen data.

The training set comprises the majority of the images, the validation set is used to monitor overfitting and tune hyperparameters during training, while the test set is kept entirely separate and only used for the final evaluation of the models.

Due to the class imbalance inherent in the dataset, we ensure the representation of minority classes in all splits. The training and validation datasets contain fewer images from minority classes compared to majority ones, which motivates the generation and incorporation of synthetic images to mitigate the class imbalance. 

All images are subjected to a standardized preprocessing pipeline prior to training. First, images are resized to a consistent resolution compatible with the input size requirements of the classifiers, specifically ResNet-50 and CLIP. Next, normalization is applied to scale pixel intensity values, enhancing the stability and convergence of the models during training.

To further improve model robustness, data augmentation techniques is applied on the training set, including random rotations, flips, and color jittering, simulating variations that occur in real microscopy images. No augmentation is applied to the validation and test sets to ensure unbiased evaluation.

\subsection{Evaluation metrics}
To evaluate the quality of the synthetic images generated during the experiments, the Fréchet Inception Distance (FID) is employed. FID measures the similarity between the distributions of real and synthetic images by comparing the statistics of features extracted from a pretrained Inception network \cite{wu2025pragmatic}. Lower FID scores indicate higher fidelity and diversity of the generated images, reflecting their closeness to real data.

\begin{table*}[ht]
    \centering
    \caption{Classification performance of ResNet-50 and CLIP models trained on 18,365 real images from 15 highly imbalanced blood cell classes. Results are shown for experiments using only real images and for experiments combining real and 5,000 synthetic images per class. The dataset is split into 60\% training, 20\% validation, and 20\% test sets. Metrics are averaged over 5 cross-validation folds.}

    \vspace{-0.3cm}
    \tabcolsep=0.03\textwidth
    \normalsize
    \begin{tabular}{lcccc}
        \toprule
        \textbf{Model (Split)} & \textbf{Dataset} & \textbf{Accuracy} & \textbf{F1 macro} & \textbf{AUC} \\
        \midrule
        ResNet-50 (Validation) &  & 0.40 \scriptsize$\pm$0.07 & 0.08 \scriptsize$\pm$0.03 & 0.57 \scriptsize$\pm$0.07 \\
        ResNet-50 (Test)       & Real only & 0.27 \scriptsize$\pm$0.10 & 0.07 \scriptsize$\pm$0.01 & 0.56 \scriptsize$\pm$0.03 \\
        CLIP (Validation)      &  & 0.78 \scriptsize$\pm$0.01 & 0.27 \scriptsize$\pm$0.05 & 0.88 \scriptsize$\pm$0.02 \\
        CLIP (Test)            &  & 0.62 \scriptsize$\pm$0.03 & 0.23 \scriptsize$\pm$0.02 & 0.83 \scriptsize$\pm$0.02 \\
        \midrule
        ResNet-50 (Validation) &  & 0.89 \scriptsize$\pm$0.01 & 0.90 \scriptsize$\pm$0.01 & 0.99 \scriptsize$\pm$0.00 \\
        ResNet-50 (Test)       & Real + Synthetic & 0.78 \scriptsize$\pm$0.01 & 0.80 \scriptsize$\pm$0.01 & 0.99 \scriptsize$\pm$0.00 \\
        CLIP (Validation)      &  & 0.85 \scriptsize$\pm$0.02 & 0.85 \scriptsize$\pm$0.02 & 0.99 \scriptsize$\pm$0.00 \\
        CLIP (Test)            &  & 0.77 \scriptsize$\pm$0.01 & 0.78 \scriptsize$\pm$0.01 & 0.98 \scriptsize$\pm$0.00 \\
        \bottomrule
    \end{tabular}
    \label{tab:resnet_clip_real2}
\end{table*}

For assessing the performance of the classification models, several standard metrics are used to capture different aspects of their predictive ability. The overall accuracy is computed as the proportion of correctly classified samples among all predictions, providing a general measure of the model correctness. To better handle class imbalance and provide a more nuanced evaluation, the macro-averaged F1-score is calculated. This metric is the harmonic mean of precision and recall averaged equally across all classes, thus reflecting the balance between false positives and false negatives on a per-class basis. Additionally, the Area Under the Receiver Operating Characteristic Curve (AUC-ROC) is computed in a one-vs-all scheme for each class. Unlike accuracy and macro F1-score, which rely on the final class prediction obtained via the argmax of the softmax outputs, the AUC-ROC metric works directly with the predicted class probabilities. This allows for evaluating the model's ability to distinguish between classes across different decision thresholds, offering deeper insight into its discriminative power beyond the final decision.

Together, these metrics offer a comprehensive evaluation framework to measure both image generation quality and classifier effectiveness across the multi-class setting.

\subsection{Implementation details}

The models are optimized using the AdamW optimizer, which combines adaptive moment estimation with decoupled weight decay (set to $1\mathrm{e}{-8}$) to promote better generalization. A cosine learning rate scheduler with warm-up is employed to adjust the learning rate dynamically across epochs. The learning rate is initially set to $1\mathrm{e}{-4}$, with a minimum value of $1\mathrm{e}{-8}$, and the warm-up phase lasted for 30 epochs. Each model is trained for 100 epochs, using a batch size of 64 for training and 1 for evaluation.

\subsection{Quantitative results comparison}
\label{sec:Quantitative results comparison}
\subsubsection{Classification performance on real imbalanced data}
Subsequently, an analysis of the metrics of the two classifiers utilized is performed. The initial experiment is conducted employing real images solely to determine our research question, as synthetic images can improve the performance of classification models and support medical diagnosis.

This outcome is expected under conditions of high class imbalance. The failure of ResNet-50 to consider minority classes has the effect of creating a false impression of high performance. As illustrated in Table~\ref{tab:resnet_clip_real2}, the accuracy achieved is 27\%, which is four times higher than the performance of random guessing. However, examining the confusion matrix, provided in the Supplementary Figure S3, reveals that it only correctly classifies samples exclusively from the Segmented Neutrophil, Typical Lymphocyte, and Monocyte classes, which are three of the most represented classes in our dataset.

In the scenario of imbalanced data, it is crucial to avoid reliance on accuracy metrics and to employ alternative metrics that are more reliable in such circumstances. One such metric is F1-macro score. This metric treats all classes equally, so it penalizes the model if it only classifies the majority classes correctly. The F1-Macro score of 0.075 indicates that the model is nearly random, rendering it incapable of differentiating between distinct cell classes and making it obviously insufficient for clinical applications.

Concurrently, CLIP demonstrated superior performance. As illustrated in Table~\ref{tab:resnet_clip_real2}, the model achieves accuracy of 62\%, along with an F1-macro score of 0.23, which is approximately three times more accurate than the ResNet-50 model. The findings indicate that CLIP exhibits superior performance in scenarios characterized by significant class imbalance. This is due to the fact that, in contrast to receiving only images, CLIP receives a pair consisting of an image and the prompts for each class. This capability enables the model to meticulously analyze the morphological characteristics delineated in the prompts, thereby facilitating a more precise classification of the given images. Further insights into CLIP’s classification performance are found in the confusion matrix shown in Supplementary Figure S4.

A notable observation is that both classifiers exhibit significant overfitting, as evidenced by the substantial discrepancy between the validation and test metrics. This could be because training for 100 epochs is excessive in this context. This causes the model to memorize the training data instead of generalizing them.




\subsubsection{Classification performance on only synthetic data}

Both classifiers are trained exclusively using 3,000 synthetic images for each class. This experiment is of significant value; it enables a comparison of the two classifiers with a perfectly balanced dataset. Conversely, it facilitates the evaluation of the generated synthetic images, as optimal classification outcomes signify that the classes possess distinctive and discernible morphological characteristics.

The results using synthetic images exclusively have been successful, demonstrating the best results observed. ResNet-50 achieves accuracy and F1-macro score of 0.82, while the AUC is close to 0.99, indicating
excellent class differentiation. However, the CLIP model performed slightly worse, suggesting that it does not interpret
the cellular morphological details in the synthetic images as well as ResNet-50. The detailed results of this experiment
are shown in Table S1.

\subsubsection{Classification performance on real and synthetic data}
In this setting, we evaluate the impact of combining real and synthetic images for training, which represents the main goal of this study. The experimental setup and results are presented in \ref{sec:Accuracy scaling by number of synthetic data}, where we analyze how classification accuracy scales as the number of synthetic images per class increases. 

Although it may appear counter-intuitive that training solely on synthetic data yields better performance than combining real and synthetic images, as observed when comparing Table~\ref{tab:resnet_clip_real2} and S1, this result can be partially explained by the limitations of our dataset. Due to the extremely low number of real images in certain classes, some with as few as 16 instances, it was not feasible to perform K-fold cross-validation with a test set composed exclusively of real data while maintaining statistical robustness. A test set formed only by real images would contain very few samples per class (e.g., 3–4), making it unreliable for evaluating performance metrics such as accuracy or F1-score. Therefore, we constructed the test sets by combining both real and synthetic images, which was necessary to ensure all classes were represented. However, this setup introduces a domain bias that may favor models trained entirely on synthetic data, as the test distribution partially overlaps with the training domain. This constraint reflects a trade-off between experimental rigor and the practical limitations of working with small biomedical datasets.

\subsection{Ablation study}
\label{sec:Ablation study}
\subsubsection{Varying the number of few-shot real images for synthetic image generation}
\label{sec:few-shot ablation study}
\begin{figure}[t!]
    \centering
    \setlength{\tabcolsep}{5pt}
    \renewcommand{\arraystretch}{1.2}
    \begin{tabular}{>{\centering\arraybackslash}m{1.1cm} c c c c}
        \shortstack{Real \\ images} &
        \includegraphics[width=0.17\columnwidth]{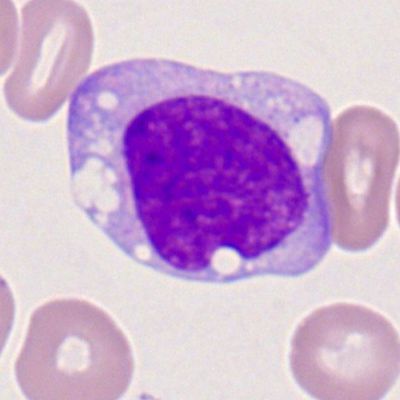} & 
        \includegraphics[width=0.17\columnwidth]{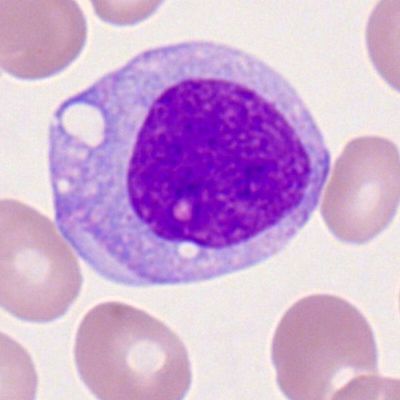} & 
        \includegraphics[width=0.17\columnwidth]{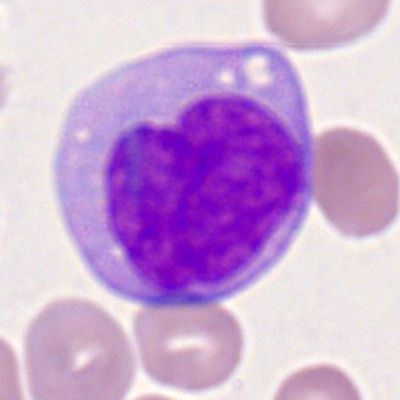} & 
        \includegraphics[width=0.17\columnwidth]{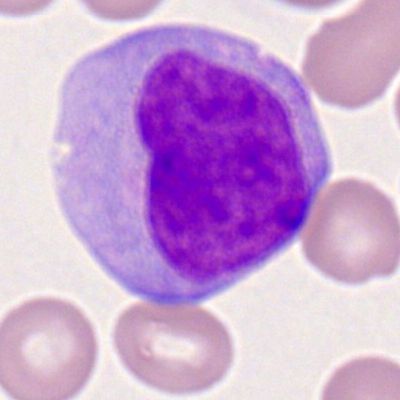} \\
        
        1-shot & 
        \includegraphics[width=0.17\columnwidth]{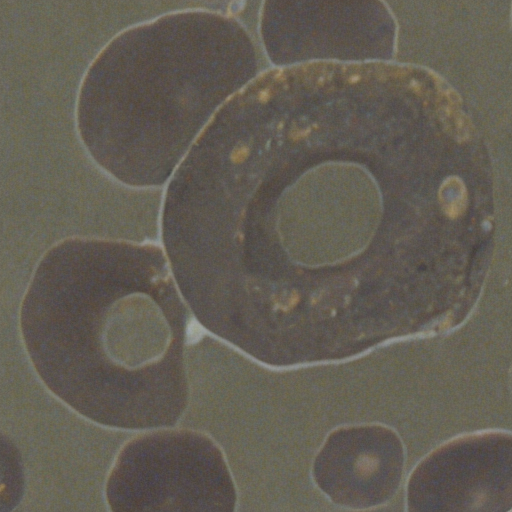} & 
        \includegraphics[width=0.17\columnwidth]{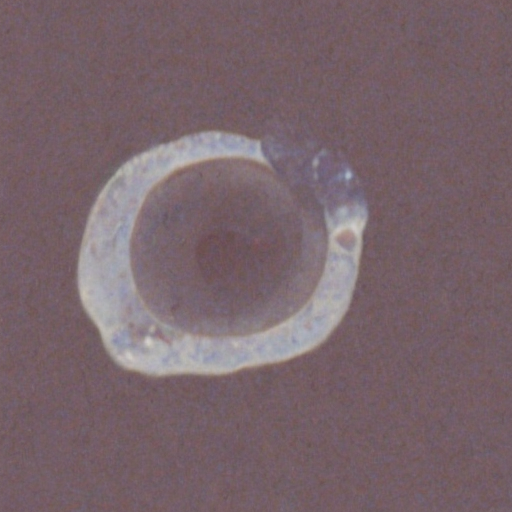} & 
        \includegraphics[width=0.17\columnwidth]{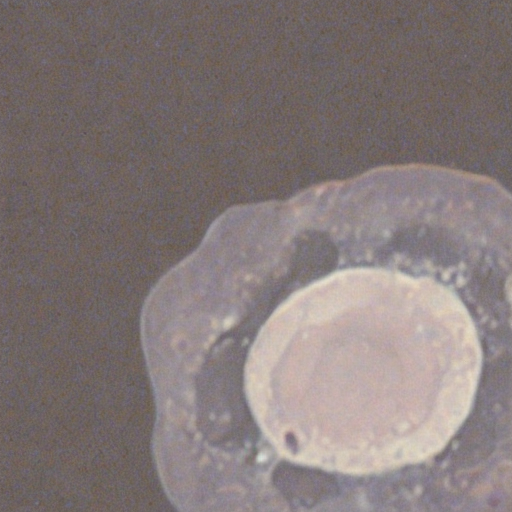} & 
        \includegraphics[width=0.17\columnwidth]{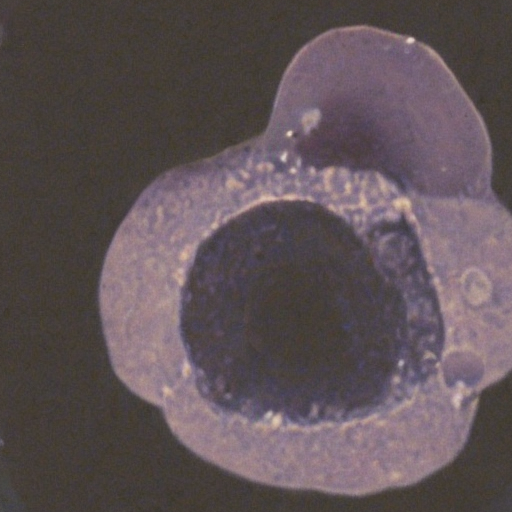} \\
        
        4-shot & 
        \includegraphics[width=0.17\columnwidth]{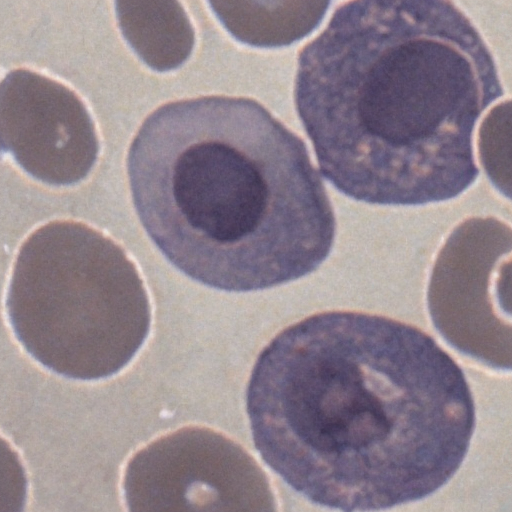} & 
        \includegraphics[width=0.17\columnwidth]{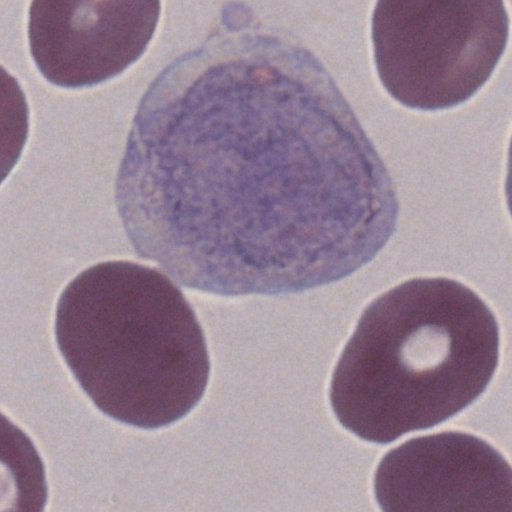} & 
        \includegraphics[width=0.17\columnwidth]{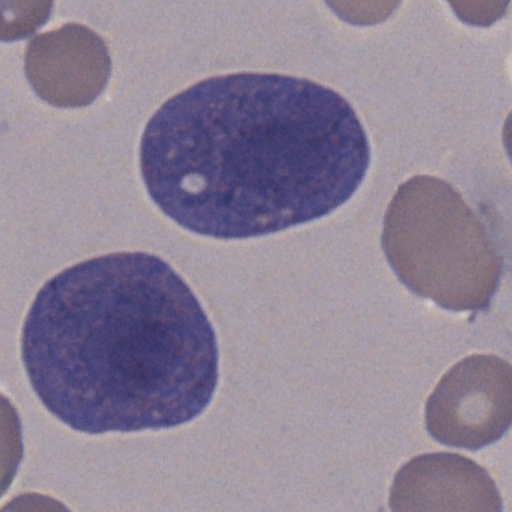} & 
        \includegraphics[width=0.17\columnwidth]{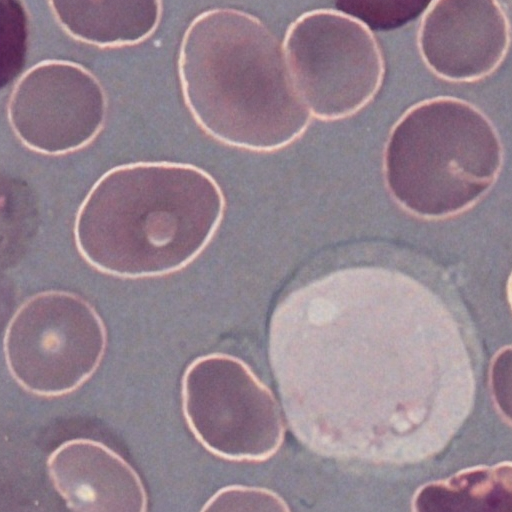} \\
        
        8-shot & 
        \includegraphics[width=0.17\columnwidth]{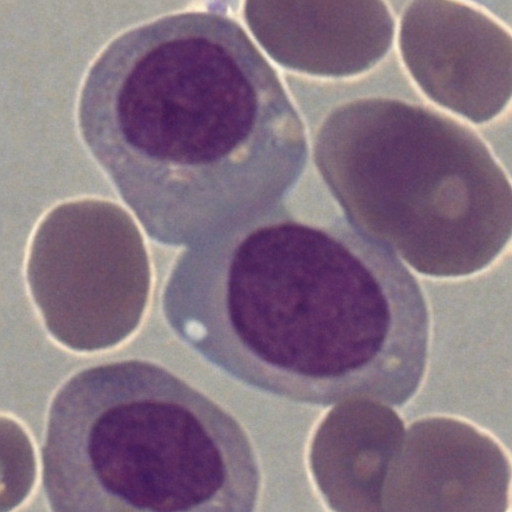} & 
        \includegraphics[width=0.17\columnwidth]{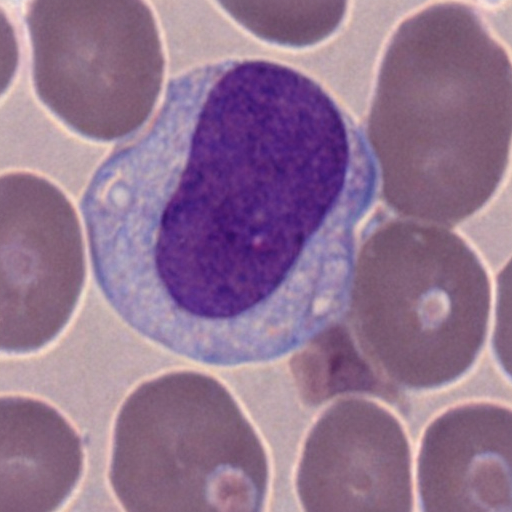} & 
        \includegraphics[width=0.17\columnwidth]{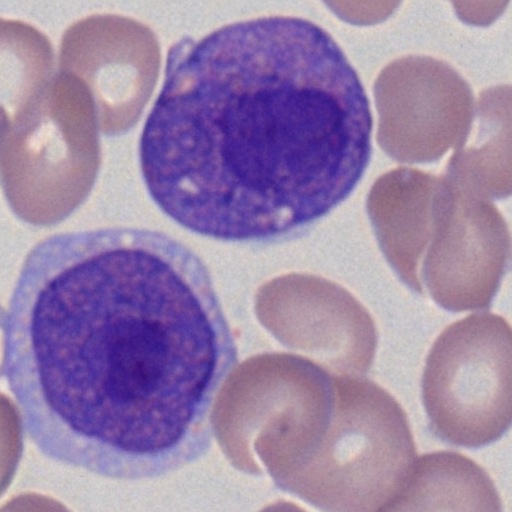} & 
        \includegraphics[width=0.17\columnwidth]{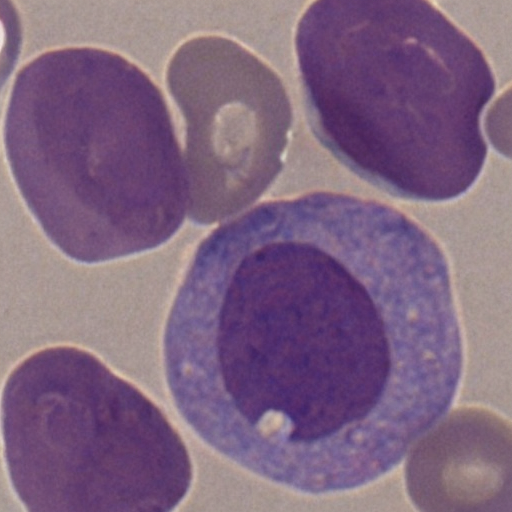} \\
        
        16-shot & 
        \includegraphics[width=0.17\columnwidth]{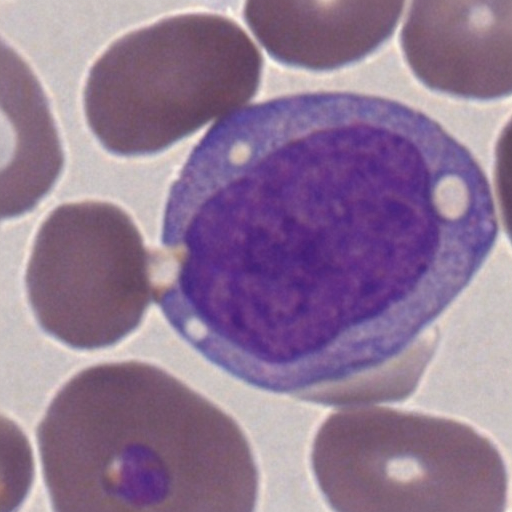} & 
        \includegraphics[width=0.17\columnwidth]{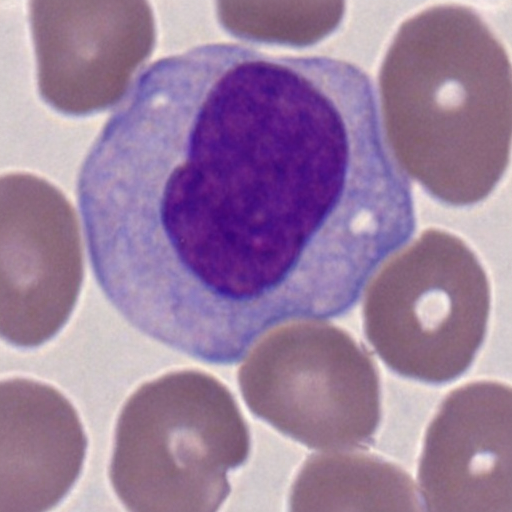} & 
        \includegraphics[width=0.17\columnwidth]{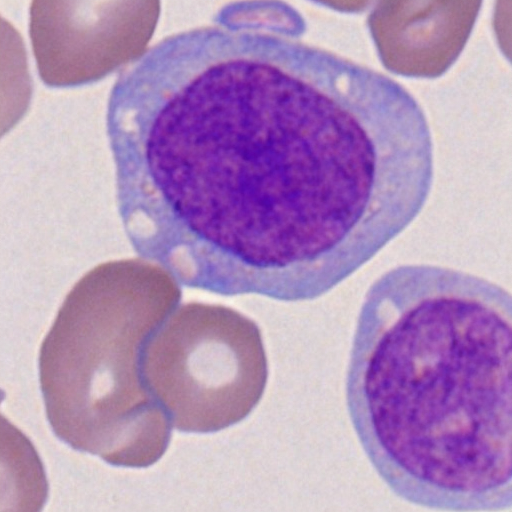} & 
        \includegraphics[width=0.17\columnwidth]{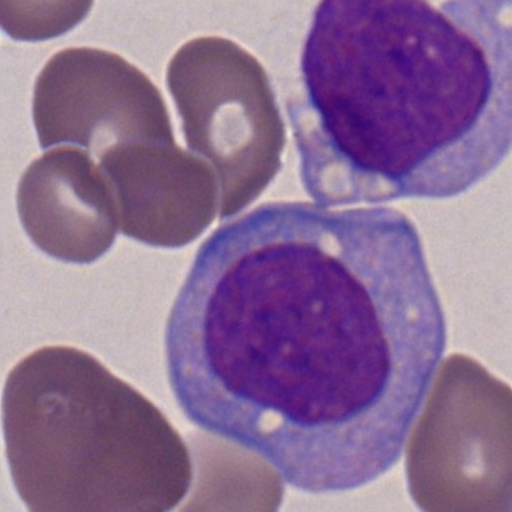} \\
    \end{tabular}
    \caption{Real and synthetic monoblast images generated by CytoDiff with increasing number of real few-shot images.}
    \label{fig:n_shot_monoblast}
\end{figure}

Initially, an evaluation of the synthetic images generated by CytoDiff model with the implementation of LoRA weights in the Attention Layers has been conducted. The presence of morphological details in the images is a key to optimal training of the classifiers. This is exemplified in Figure~\ref{fig:n_shot_monoblast}, which shows five rows, each containing four images. The initial row displays real Monoblast images from the Munich AML Morphology Dataset \cite{matek2019dataset}. The subsequent four rows present synthetic images generated by our CytoDiff model. In the first instance, the model is trained with a single real image, the second instance involved four images, the third instance involved eight images, and the final instance involved sixteen images. A clear progression is observed as the number of real few-shot images used to train the model increases, resulting in enhanced realism and similarity to real images.  

In the first row, the cell is barely recognizable, with only basic features such as the cytoplasm and nucleus being identifiable. In the second row, a clear distinction between red blood cells and monoblasts is already apparent, as is typically observed in a peripheral blood smear.  The final two rows exhibit a significantly higher degree of similarity to the real images, making them almost indistinguishable. Unique morphological characteristics of the monoblast is observed, such as a rounded or slightly oval nucleus, a cytoplasm that appears more bluish than the nucleus, and the presence of fine cytoplasmic vacuoles.

The supplementary material contains further examples of other cell types that follow the same structure (see, e.g., Figure S5 and Figure S6). It is observed that most synthetic images appear darker than the real ones. This difference may be attributed to variations in illumination, staining, or other acquisition-related factors, specified in the customized text prompts.

The image generation process has been successful for most cell classes; however, certain classes show less satisfactory results, as illustrated in the comparison between real and synthetic images in Figure~\ref{fig:comparison_real_synthetic}. A notable example of this issue is the Promyelocyte Bilobed. The generated images presented  in Figure S7, are characterized by a lack of sharpness, which complicates their recognition. This may be attributed to two primary factors. The primary and most apparent aspect is the quality of the real images. Despite the implementation of a manual selection process to identify the best real images to train CytoDiff, there are classes, such as Promyelocyte Bilobed, where the quality of these images is considerably lower in comparison to other classes.  This phenomenon is explicable, as this particular cell type is uncommon. A promyelocyte with a bilobed nucleus is typically observed in patients diagnosed with acute myeloid leukemia.

Another potential factor is that the CytoDiff model may not be trained on a specific type of image, such as a cellular anomaly. Instead, it is possible that the model has been trained in more common classes as neutrophils or lymphocytes.

To quantitatively assess the quality and realism of the images generated by the models, the Fréchet Inception Distance (FID) is employed. A notable limitation of employing FID with our dataset is the necessity of a minimum of 1,000 real images and 1,000 synthetic images to ensure reliable results. In the scenario of the Munich AML Morphology Dataset \cite{matek2019dataset}, the number of real samples available is limited to four classes: Segmented Neutrophil, Monocyte, Myeloblast, and Typical Lymphocyte. As indicated in supplementary Figure S8, the range of FID scores obtained is approximately 50 to 90. The absence of an objective scale to evaluate the quality of these results necessitates a comparison with other studies. The images generated in DataDream~\cite{kim2024datadream} predominantly exhibit FID values ranging from 40 to 60. Although it is true that the images generated by CytoDiff show higher FID scores. It is important to note that our method involves the generation of microscopic-level blood cell images, rather than simpler visual structures such as airplanes or cats.

\label{sec:Accuracy scaling by number of synthetic data}

\subsubsection{Accuracy scaling by number of synthetic data}

Thereafter, different experiments are conducted  by progressively adding increments of 100 synthetic images per class to the existing real images, until reaching a total of 1,000 per class. The results, summarized in Table S2, present a notable enhancement in all metrics using ResNet-50. As more images are added, the classifier starts to correctly classify the minority classes, thereby aligning the accuracy and F1-macro score.

The results are very promising. To illustrate, in the experiment involving 700 synthetic images per class, both accuracy and F1-Macro exhibit a value of 0.60, while the Area Under the Curve (AUC) attains a value of 0.93.

In contrast to other metrics, AUC functions with probabilities rather than with a final decision. The model's high AUC value, when compared to the other metrics, suggests that it is learning to distinguish the classes effectively in the representation space. However, its final prediction is not entirely accurate.

\begin{figure}[t!]
    \centering
    \includegraphics[width=\linewidth]{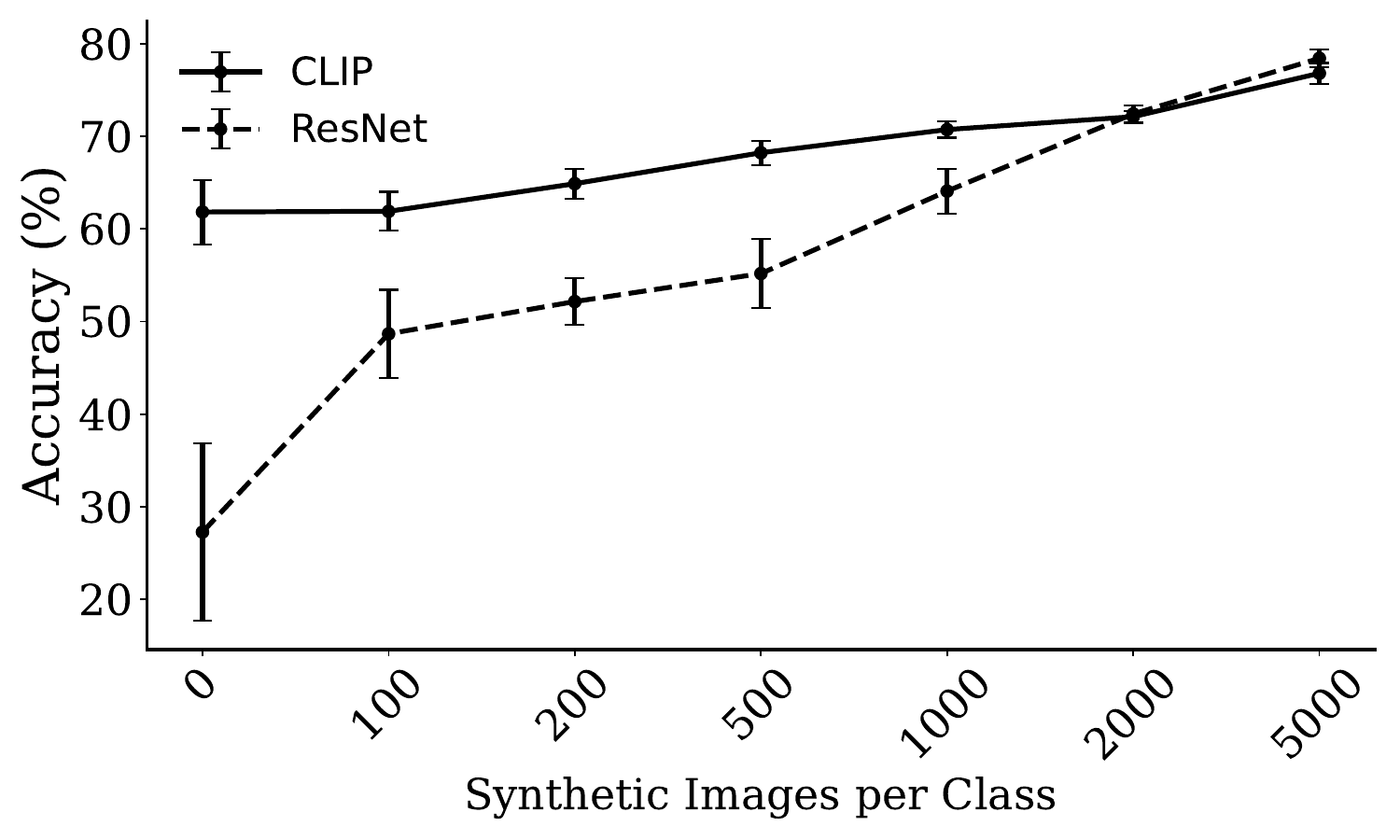}
    \caption{Classification accuracy increases when the real dataset images is augmented with additional synthetic images per class. Each point shows the mean accuracy across 5 cross-validation folds, with error bars representing the standard deviation.}
    \label{fig:clip_accuracy_evolution_col}
\end{figure}

Given the continuous enhancement of the metrics, the experiment series is expanded by incorporating an additional 2,000 and 3,000 synthetic images per class. A thorough examination of the confusion matrix from the previous experiment with ResNet-50, provided in Figure S9 in the supplementary material, reveals that the majority of the samples from all classes have been classified with a high degree of accuracy.

The same experiments are conducted using CLIP model. As shown in Table S3 in the supplementary material, the results indicate that the classifier demonstrates a reduced capacity for enhancement when generating synthetic images. While the accuracy of ResNet-50 increases from 27\% to 75\%, the accuracy of CLIP rises from 61\% to 73\%. In both classifiers, the overfitting is reduced as more images were used, indicating that the larger the dataset, the more epochs are required.

The primary factor contributing to the enhancement in metrics is the gradual balancing of the dataset over time. When utilizing exclusively real images, the four majority classes constitute 95\% of the dataset. However, when 3,000 synthetic images per class are incorporated, these four classes constitute slightly over 46\% of the dataset. This modification balances the metrics of both classifiers, leading to a substantial enhancement in ResNet-50 and a minor improvement in CLIP. As observed in the confusion matrix provided in the supplementary material S10, CLIP can accurately classify the majority of samples, such as ResNet-50.

Given the persistent upward trend in metrics without any discernible indications of stabilization, two concluding experiments are conducted, using 4,000 and 5,000 synthetic images per class. The Figure \ref{fig:clip_accuracy_evolution_col} illustrates the continued increase in metrics, suggesting a discernible advantage in relation to the number of images used.






\section{Conclusion}
Our work has demonstrated that the generation of synthetic images using CytoDiff, a cytomorphology stable diffusion model, fine-tuned with LoRA weights, is a tool to be considered in the biomedical field. The quality of synthetic images exhibits cellular morphological features that closely resemble real ones, especially when high-quality real images are available.

From the white blood cells classification, it is stated that the inclusion of synthetic data generated by CytoDiff has led to a clear improvement in the metrics, thereby supporting the research question: synthetic images can improve the performance of classification models and support medical diagnosis.

It has been demonstrated that, on imbalanced datasets, CLIP can understand morphological details of all classes better than ResNet-50 and classify minority classes. However, when synthetic data is included, ResNet-50 showed slightly better results. This is because the dataset is larger and the classes are more balanced.
Ultimately, ResNet-50 has been shown to be more effective in interpreting and classifying synthetic images.

CytoDiff has the potential to facilitate the unlimited and realistic generation of biomedical data, thereby addressing privacy-related legal concerns and logistical challenges. It has the potential to expedite and improve clinical research, particularly for rare classes of cytomorphology and hematology.

In conclusion, this work establishes a substantial foundation for the consideration of synthetic images as a complementary tool in the biomedical field. In the long term, synthetic images could potentially replace real images in specific biomedical applications, helping to advance the fields of medical research and diagnosis.

\section*{Acknowledgements}
J.C.B. acknowledges Helmholtz Munich HPC compute resources for all experiments. J.C.B. and R.M.U. acknowledge feedback of Muhammed Furkan Dasdelen, Helmholtz Munich for guiding the text prompts of the single white blood cell generation task. C.M. acknowledges funding from the European Research Council (ERC) under the European Union's Horizon 2020 research and innovation program (Grant Agreement No. 866411 \& 101113551 \& 101213822) and support from the
Hightech Agenda Bayern.

{
    \small
    \bibliographystyle{ieeenat_fullname}
    \bibliography{refs}

\begin{thebibliography}{23}
\providecommand{\natexlab}[1]{#1}
\providecommand{\url}[1]{\texttt{#1}}
\expandafter\ifx\csname urlstyle\endcsname\relax
  \providecommand{\doi}[1]{doi: #1}\else
  \providecommand{\doi}{doi: \begingroup \urlstyle{rm}\Url}\fi

\bibitem[Banerjee et~al.(2023)Banerjee, Taroni, Allaway, Prasad, Guinney, and Greene]{banerjee2023machine}
Jineta Banerjee, Jaclyn~N Taroni, Robert~J Allaway, Deepashree~Venkatesh Prasad, Justin Guinney, and Casey Greene.
\newblock Machine learning in rare disease.
\newblock \emph{Nature methods}, 20\penalty0 (6):\penalty0 803--814, 2023.

\bibitem[Betker et~al.(2023)Betker, Goh, Jing, Brooks, Wang, Li, Ouyang, Zhuang, Lee, Guo, et~al.]{betker2023improving}
James Betker, Gabriel Goh, Li Jing, Tim Brooks, Jianfeng Wang, Linjie Li, Long Ouyang, Juntang Zhuang, Joyce Lee, Yufei Guo, et~al.
\newblock Improving image generation with better captions.
\newblock \emph{Computer Science}, 2023.

\bibitem[Bhosale et~al.(2025)Bhosale, Wasi, Zhai, Tian, Border, Xi, Sarder, Yuan, Doermann, and Gong]{bhosale2025pathdiff}
Mahesh Bhosale, Abdul Wasi, Yuanhao Zhai, Yunjie Tian, Samuel Border, Nan Xi, Pinaki Sarder, Junsong Yuan, David Doermann, and Xuan Gong.
\newblock Pathdiff: Histopathology image synthesis with unpaired text and mask conditions.
\newblock \emph{arXiv preprint arXiv:2506.23440}, 2025.

\bibitem[Choi et~al.(2024)Choi, Park, Park, Cho, No, and Ryu]{choi2024simple}
Joo~Young Choi, Jaesung~R Park, Inkyu Park, Jaewoong Cho, Albert No, and Ernest~K Ryu.
\newblock Simple drop-in lora conditioning on attention layers will improve your diffusion model.
\newblock \emph{arXiv preprint arXiv:2405.03958}, 2024.

\bibitem[Coyner et~al.(2022)Coyner, Chen, Chang, Singh, Ostmo, Chan, Chiang, Kalpathy-Cramer, Campbell, Imaging, in~Retinopathy~of Prematurity~Consortium, et~al.]{coyner2022synthetic}
Aaron~S Coyner, Jimmy~S Chen, Ken Chang, Praveer Singh, Susan Ostmo, RV~Paul Chan, Michael~F Chiang, Jayashree Kalpathy-Cramer, J~Peter Campbell, Imaging, Informatics in~Retinopathy~of Prematurity~Consortium, et~al.
\newblock Synthetic medical images for robust, privacy-preserving training of artificial intelligence: application to retinopathy of prematurity diagnosis.
\newblock \emph{Ophthalmology Science}, 2\penalty0 (2):\penalty0 100126, 2022.

\bibitem[Goodfellow et~al.(2014)Goodfellow, Pouget-Abadie, Mirza, Xu, Warde-Farley, Ozair, Courville, and Bengio]{goodfellow2014generative}
Ian~J Goodfellow, Jean Pouget-Abadie, Mehdi Mirza, Bing Xu, David Warde-Farley, Sherjil Ozair, Aaron Courville, and Yoshua Bengio.
\newblock Generative adversarial nets.
\newblock \emph{Advances in neural information processing systems}, 27, 2014.

\bibitem[He et~al.(2016)He, Zhang, Ren, and Sun]{he2016deep}
Kaiming He, Xiangyu Zhang, Shaoqing Ren, and Jian Sun.
\newblock Deep residual learning for image recognition.
\newblock In \emph{Proceedings of the IEEE Conference on Computer Vision and Pattern Recognition}, pages 770--778, 2016.

\bibitem[Kim et~al.(2024)Kim, Bader, Alaniz, Schmid, and Akata]{kim2024datadream}
Jae~Myung Kim, Jessica Bader, Stephan Alaniz, Cordelia Schmid, and Zeynep Akata.
\newblock Datadream: Few-shot guided dataset generation.
\newblock In \emph{European Conference on Computer Vision}, pages 252--268. Springer, 2024.

\bibitem[Matek et~al.(2019)Matek, Schwarz, Marr, and Spiekermann]{matek2019dataset}
Christian Matek, Stefan Schwarz, Claudia Marr, and Klaus Spiekermann.
\newblock A single-cell morphological dataset of leukocytes from aml patients and non-malignant controls.
\newblock The Cancer Imaging Archive, 2019.
\newblock [Data set].

\bibitem[Pezoulas et~al.(2024)Pezoulas, Zaridis, Mylona, Androutsos, Apostolidis, Tachos, and Fotiadis]{pezoulas2024synthetic}
Vasileios~C Pezoulas, Dimitrios~I Zaridis, Eugenia Mylona, Christos Androutsos, Kosmas Apostolidis, Nikolaos~S Tachos, and Dimitrios~I Fotiadis.
\newblock Synthetic data generation methods in healthcare: A review on open-source tools and methods.
\newblock \emph{Computational and structural biotechnology journal}, 2024.

\bibitem[Podell et~al.(2023)Podell, English, Lacey, Blattmann, Dockhorn, M{\"u}ller, Penna, and Rombach]{podell2023sdxl}
Dustin Podell, Zion English, Kyle Lacey, Andreas Blattmann, Tim Dockhorn, Jonas M{\"u}ller, Joe Penna, and Robin Rombach.
\newblock Sdxl: Improving latent diffusion models for high-resolution image synthesis.
\newblock In \emph{ICLR}, 2023.

\bibitem[Radford et~al.(2021)Radford, Kim, Hallacy, Ramesh, Goh, et~al.]{radford2021learning}
Alec Radford, Jong~Wook Kim, Chris Hallacy, Aditya Ramesh, Gabriel Goh, et~al.
\newblock Learning transferable visual models from natural language supervision.
\newblock 2021.

\bibitem[Rieke et~al.(2020)Rieke, Hancox, Li, Milletari, Roth, Albarqouni, Bakas, Galtier, Landman, Maier-Hein, et~al.]{rieke2020future}
Nicola Rieke, Jonny Hancox, Wenqi Li, Fausto Milletari, Holger~R Roth, Shadi Albarqouni, Spyridon Bakas, Mathieu~N Galtier, Bennett~A Landman, Klaus Maier-Hein, et~al.
\newblock The future of digital health with federated learning.
\newblock \emph{NPJ digital medicine}, 3\penalty0 (1):\penalty0 119, 2020.

\bibitem[Rombach et~al.(2022)Rombach, Blattmann, Lorenz, Esser, and Ommer]{rombach2022high}
Robin Rombach, Andreas Blattmann, Dominik Lorenz, Patrick Esser, and Bj{\"o}rn Ommer.
\newblock High-resolution image synthesis with latent diffusion models.
\newblock In \emph{CVPR}, 2022.

\bibitem[Ruiz et~al.(2023)Ruiz, Li, Jampani, Pritch, Rubinstein, and Aberman]{ruiz2023dreambooth}
Nataniel Ruiz, Yuanzhen Li, Varun Jampani, Yael Pritch, Michael Rubinstein, and Kfir Aberman.
\newblock Dreambooth: Fine tuning text-to-image diffusion models for subject-driven generation.
\newblock In \emph{CVPR}, 2023.

\bibitem[Saharia et~al.(2022)Saharia, Chan, Saxena, Li, Whang, Denton, Ghasemipour, Gontijo~Lopes, Karagol~Ayan, Salimans, et~al.]{saharia2022photorealistic}
Chitwan Saharia, William Chan, Saurabh Saxena, Lala Li, Jay Whang, Emily~L Denton, Kamyar Ghasemipour, Raphael Gontijo~Lopes, Burcu Karagol~Ayan, Tim Salimans, et~al.
\newblock Photorealistic text-to-image diffusion models with deep language understanding.
\newblock In \emph{NeurIPS}, 2022.

\bibitem[Sheng et~al.(2025)Sheng, Keane, Tham, and Wong]{sheng2025synthetic}
Bin Sheng, Pearse~A Keane, Yih-Chung Tham, and Tien~Yin Wong.
\newblock Synthetic data boosts medical foundation models.
\newblock \emph{Nature Biomedical Engineering}, 9\penalty0 (4):\penalty0 443--444, 2025.

\bibitem[Sun et~al.(2025)Sun, Tan, Gu, He, Chen, Pang, and Yan]{sun2025data}
Yuqi Sun, Weimin Tan, Zhuoyao Gu, Ruian He, Siyuan Chen, Miao Pang, and Bo Yan.
\newblock A data-efficient strategy for building high-performing medical foundation models.
\newblock \emph{Nature Biomedical Engineering}, pages 1--13, 2025.

\bibitem[Susladkar et~al.(2025)Susladkar, Deshmukh, Tur, Durak, and Bagci]{susladkar2025victr}
Onkar Susladkar, Gayatri Deshmukh, Yalcin Tur, Gorkhem Durak, and Ulas Bagci.
\newblock Victr: Vital consistency transfer for pathology aware image synthesis.
\newblock \emph{arXiv preprint arXiv:2505.04963}, 2025.

\bibitem[Umer et~al.(2023)Umer, Gruber, Shetab~Boushehri, Metak, and Marr]{umer2023imbalanced}
Rao~Muhammad Umer, Armin Gruber, Sayedali Shetab~Boushehri, Christian Metak, and Carsten Marr.
\newblock Imbalanced domain generalization for robust single cell classification in hematological cytomorphology.
\newblock In \emph{Proceedings of the 11th International Conference on Learning Representations (ICLR) Workshops}, 2023.

\bibitem[Walter et~al.(2022)Walter, Pohlkamp, Meggendorfer, Nadarajah, Kern, Haferlach, and Haferlach]{walter2022artificial}
Wencke Walter, Christian Pohlkamp, Manja Meggendorfer, Niroshan Nadarajah, Wolfgang Kern, Claudia Haferlach, and Torsten Haferlach.
\newblock Artificial intelligence in hematological diagnostics: Game changer or gadget?
\newblock \emph{Blood Reviews}, page 101019, 2022.

\bibitem[Wu et~al.(2025)Wu, Liu, Yilmaz, Konermann, Walter, and Stegmaier]{wu2025pragmatic}
Yuli Wu, Fucheng Liu, R{\"u}veyda Yilmaz, Henning Konermann, Peter Walter, and Johannes Stegmaier.
\newblock A pragmatic note on evaluating generative models with fr$\backslash$'echet inception distance for retinal image synthesis.
\newblock \emph{arXiv preprint arXiv:2502.17160}, 2025.

\bibitem[Ye-Bin et~al.(2025)Ye-Bin, Hyeon-Woo, Choi, Kim, Kwak, and Oh]{ye2023synaug}
Moon Ye-Bin, Nam Hyeon-Woo, Wonseok Choi, Nayeong Kim, Suha Kwak, and Tae-Hyun Oh.
\newblock Synaug: Exploiting synthetic data for data imbalance problems.
\newblock \emph{Pattern Recognition Letters}, 2025.

\end{thebibliography}
}

\end{document}